\documentclass[sn-mathphys,Numbered]{sn-jnl}

\usepackage{graphicx}%
\usepackage{multirow}%
\usepackage{amsmath,amssymb,amsfonts}%
\usepackage{amsthm}%
\usepackage{mathrsfs}%
\usepackage[title]{appendix}%
\usepackage{xcolor}%
\usepackage{textcomp}%
\usepackage{manyfoot}%
\usepackage{booktabs}%
\usepackage{algorithm}%
\usepackage{algorithmicx}%
\usepackage{algpseudocode}%
\usepackage{listings}%

\theoremstyle{thmstyleone}%
\theoremstyle{thmstyletwo}%

\theoremstyle{thmstylethree}%

\begin{document}

\title[Finding doctors and locations]{A Knowledge Graph-Based Search Engine for Robustly Finding Doctors and Locations in the Healthcare Domain}


\author*[1]{\fnm{Mayank} \sur{Kejriwal}}\email{kejriwal@isi.edu}

\author[2]{\fnm{Hamid} \sur{Haidarian}}\email{Hamid.Haidarian@kp.org}

\author[1]{\fnm{Min-Hsueh} \sur{Chiu}}\email{minhsueh@isi.edu}

\author[2]{\fnm{Andy} \sur{Xiang}}\email{Andy.X.Xiang@kp.org}

\author[2]{\fnm{Deep} \sur{Shrestha}}\email{Deep.X.Shrestha@kp.org}

\author[2]{\fnm{Faizan} \sur{Javed}}\email{faizan.x.javed@kp.org}

\affil[1]{\orgname{University of Southern California}, \country{USA}}

\affil[2]{\orgdiv{Kaiser Permanente Digital}, \country{USA}}



\abstract{Efficiently finding doctors and locations is an important search problem for patients in the healthcare domain, for which traditional information retrieval methods tend not to work optimally. In the last ten years, knowledge graphs (KGs) have emerged as a powerful way to combine the benefits of gleaning insights from semi-structured data using semantic modeling, natural language processing techniques like information extraction, and robust querying using structured query languages like SPARQL and Cypher. In this short paper, we present a KG-based search engine architecture for robustly finding doctors and locations in the healthcare domain. Early results demonstrate that our approach can lead to significantly higher coverage for complex queries without degrading quality. }




\maketitle

As the costs and complexity of healthcare continue to increase \cite{healthcarecomplexity}, technology can play an important role in  helping connect patients with the right providers and hospitals for conditions that they may be seeking services or treatment for. Such a search engine for \emph{finding doctors and locations} (FDL) can be of immense benefit to patients. While ordinary search engines, such as Google, can retrieve some relevant results, they lack specificity in their results that healthcare organizations can potentially achieve due to the latter having access to large quantities of data on providers and services. 
\begin{figure}
\centering
  \includegraphics[width=5.3in]{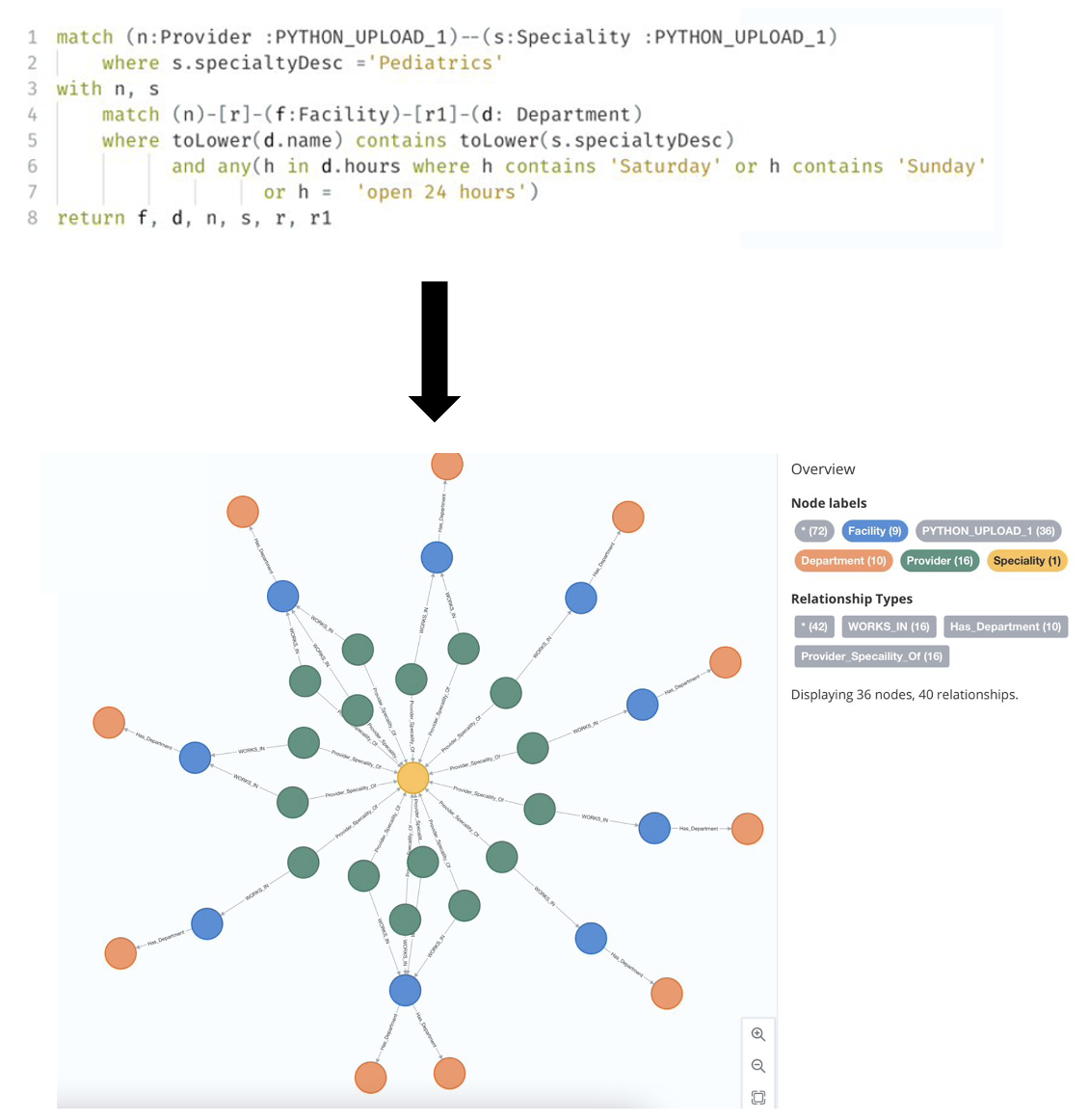}
  \caption{Example execution of a query (\emph{What pediatricians are open on the weekend near me?}) against a knowledge graph containing data on healthcare providers and services.}
  \label{example}
\end{figure}

\begin{figure*}
  \includegraphics[width=5.3in]{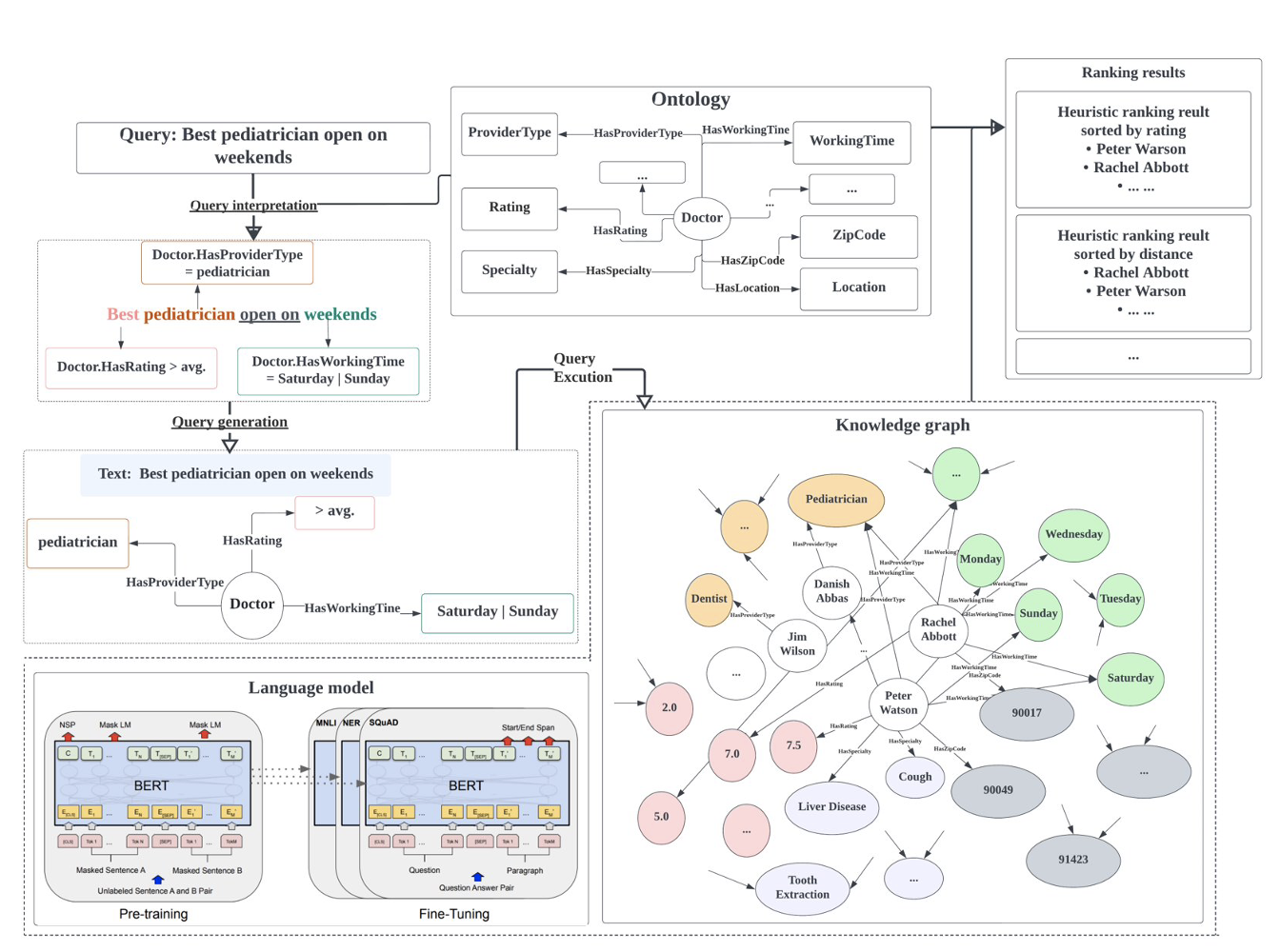}
  \caption{An architectural overview of the proposed approach}
  \label{fig:teaser}
\end{figure*}

A larger technical issue is that traditional information retrieval (IR) approaches based on matching keywords, or even more recent natural language-based approaches, may be less suitable for complex FDL queries than the structured querying approaches deployed in the database community \cite{IR1}. Consider, for example, a query such as `What pediatricians are open on the weekend near me? Order the results by closeness to my city.' Using a Lucene-based approach against a database that contains relevant options for the user, even with adjustments such as spell check and re-ranking \cite{liang2012exploiting}, we found that relevant answers were not returned in the top 10 results. One reason is that such queries require  semantic interpretation (e.g., that `open on the weekend' should be reformulated to a constraint that the hours or days of operation also include times within Saturday and Sunday) and semantic typing \cite{kejriwalST1, kejriwalST2}, but also group-by and aggregate operations that are not as well supported by natural language question answering.

This poster proposes an architecture that uses \emph{knowledge graphs} (KGs) for addressing the coverage issue in an FDL search engine based purely on keyword matching techniques \cite{kejriwalKG}.  Returning to the example in the previous paragraph, we can reformulate the question as a KG query (using a Neo4j KG implementation\footnote{Neo4j is a graph database that has recently witnessed much uptake in industry, and that has a simple graph-pattern matching query language called Cypher: \url{https://neo4j.com/}}) and execute it as shown in Figure \ref{example}. 

The overall architecture is schematized in Figure \ref{fig:teaser}. The key premise is to have a supporting \emph{semantic model} (or an ontology) rich enough to support the concepts and relations that are likely to be queried by the user \cite{healthonto}. Once such an ontology has been properly designed and implemented, and the underlying data is modeled using the ontology as a KG, a fine-tuned language model can be used to convert a natural language question into a proper query. For simpler or frequently occurring questions, query `templates' can also be used. These are pre-built queries with placeholders, which are substituted with actual data extracted from the user's question. Populating the template requires techniques like information extraction \cite{grishman2015information}, which can be difficult in domain-specific contexts \cite{kejriwalIE1,kejriwalIE2}, while the problem of directly converting a question into a Neo4j query is better suited for generative AI, being an advanced `cognitive' task \cite{kejriwalCog}. Currently, we are using a simple template-based approach, but the generative AI approach of direct question-to-query reformulation is also being considered as near-future work. Finally, as is standard in more traditional IR, either heuristics or a learning-to-rank approach (based on how much training information is available) can be used to rank the results and return them to the user \cite{liu2009learning}. In a practical implementation, both the KG-based search engine and the legacy Lucene\footnote{\url{https://lucene.apache.org/}}-based system would be used to maximize coverage. Such search approaches have  been successfully applied in other `difficult' domains, including search engines for fighting human trafficking \cite{kejriwal2017investigative}. 

To evaluate an early version of the approach, we sampled a set of almost 400 frequently used queries from Kaiser Permanente's search log, and found that at least 80 queries that were previously returning no results (from the string matching-only search engine) were now returning results using the hybrid approach that used both string matching and the knowledge graph. Because these are high-volume queries, the utility of the KG-based approach is therefore expected to be high.

In future work, we continue to integrate more sources (including news articles) to make for an even richer and more holistic search experience for the user. However, doing so will require more robust and comprehensive \emph{domain-specific} KG construction \cite{kejriwalCOVID, kejriwalMDPI}, which would include components for (domain-specific) entity resolution \cite{kejriwalER1,kejriwalER2}, noise profiling \cite{kejriwalIE3}, and possibly, incorporation of external data sources like ConceptNet \cite{kejriwalCN}. We also plan to conduct deeper experiments with `long-tail' queries that are not so frequently used, but could be critical for some users as has been found in other domains \cite{kejriwalmydig}. We will also explore in future work whether the KG-based search engine will better adapt to such queries than keyword approaches like bag-of-words, and whether advanced techniques like machine commonsense  may become  necessary for  interpreting users' true intent \cite{CSR}.


\bibliography{sn-article}

\end{document}